# An approach to melodic segmentation and classification based on filtering with the Haar-wavelet

Gissel Velarde*, Tillman Weyde** and David Meredith*

*Aalborg University, Denmark; **City University London, UK

*Correspondence:* Gissel Velarde, Aalborg University, Department of Architecture, Design and Media Technology, Sofiendalsvej 11, 9200 Aalborg SV, Denmark. Email: gv@create.aau.dk

**Abstract**

We present a novel method of classification and segmentation of melodies in symbolic representation. The method is based on filtering pitch as a signal over time with the Haar-wavelet, and we evaluate it on two tasks. The filtered signal corresponds to a single-scale signal $w_s$ from the continuous Haar wavelet transform. The melodies are first segmented using local maxima or zero-crossings of $w_s$. The segments of $w_s$ are then classified using the *k*–nearest neighbour algorithm with Euclidian and city-block distances. The method proves more effective than using unfiltered pitch signals and Gestalt-based segmentation when used to recognize the parent works of segments from Bach's *Two-Part Inventions* (BWV 772–786). When used to classify 360 Dutch folk tunes into 26 tune families, the performance of the method is comparable to the use of pitch signals, but not as good as that of string-matching methods based on multiple features.

## 1  Introduction

Melodic classification models depend strongly on melodic representation. Computational models that work on symbolic data (e.g., MIDI) usually transform the data into a suitable representation before applying any machine learning technique.

Most computational approaches for melodies use string methods, treating melodies as sequences of notes or intervals, and modelling distributions and transitions of note properties (Knopke & Jürgensen, 2009; Hillewaere, Manderick, & Conklin, 2009). Other approaches use multidimensional feature vectors to represent global properties of melodies, assigning coefficients to various musical dimensions (Ponce de Léon & Iñesta, 2004; Hillewaere, Manderick, & Conklin, 2012; van Kranenburg, 2010).

We present below a method for analysing and classifying monophonic melodies, which involves filtering symbolic representations of melodies with the Haar wavelet. We evaluate it on two classification tasks, each using a different MIDI dataset. In the first task, we use the approach to identify the parent works of segments from the parts of the fifteen *Two-Part Inventions* (BWV 772–786) by Johann Sebastian Bach (1685-1750)[1]. In the second task, the method is used to classify 360 Dutch folk songs into 26 tune families (Grijp, 2008). We compare our wavelet-based approach to the use of unfiltered pitch signals and a previous Gestalt-based model of segmentation (Cambouropoulos, 1997, 2001).

## 2 Background

### 2.1 The wavelet transform

The *wavelet transform* (WT) is a mathematical tool that was born from a multidisciplinary effort in mathematics, physics, computer science and engineering. Having developed rapidly since the second half of the 1980s, wavelets have been used for numerous applications (Daubechies, 1996; Mallat, 2009) and are today a standard tool in audio and image processing.

In the context of time-based one-dimensional (1D) signals, a wavelet is a signal that has finite energy concentrated over a short amount of time and that is zero or almost zero everywhere else. Mathematically, a wavelet is normally characterized by a total energy of 1 and an average of 0, with its energy centred around time 0 (Mallat, 2009). The WT decomposes a signal into a sum of components based on different versions of a so-called *mother wavelet* and often an additional scaling function, also called the *father wavelet*. We focus here on the mother wavelet and the coefficients that are based on shifted and scaled versions of the mother wavelet.

---

[1] We used the Musedata encodings of Bach's *Two-Part Inventions*, available at http://www.musedata.org.

Shifting refers to the position of the wavelet in time, while scaling refers to the degree of compression of the wavelet shape on the time axis, along with a normalization factor to maintain an energy of 1 (Antoine, 1999; Daubechies, 1996). The scaled and shifted versions of the wavelet are weighted by coefficients, determined by the inner product with the wavelet, so that they add up to the original signal. The wavelet transformation can also be viewed as using a filter-bank, where the coefficients at each scale correspond to a different band-pass filter that emphasises a specific scale in the signal (see Farge, 1992, pp. 449–450).

The WT is similar to the Fourier transform, with Fourier frequency corresponding to the inverse scale in wavelets. The sine and cosine functions used in Fourier analysis are periodic signals, so that the Fourier components are not localized in time within the signal being analysed. Wavelets, by contrast, have localized energy and use several shifted and scaled versions, so that wavelet coefficients become more localized in time when the scale decreases, at the expense of scale resolution. Wavelet analysis offers a trade-off between better time resolution for small scales, corresponding to high frequencies, and better scale resolution for large scales, corresponding to low frequencies (Antoine, 1999; Farge, 1992; Torrence & Compo, 1998).

There are different types of wavelets with different properties and the choice of wavelet to analyse a signal depends on the type of the signal and the features that are relevant to the analysis. There are two main forms of the WT, the *continuous wavelet transform* (CWT) and the *discrete wavelet transform* (DWT), and the two different forms tend to be used for different purposes. The CWT is mostly used for signal analysis (i.e., pattern identification or feature detection), while the DWT is used for compression and reconstruction (Antoine, 1999; Mallat, 2009). Our method is based on the CWT, which will be described below.

In audio music information retrieval (MIR), both the continuous and discrete WT have been applied extensively in tasks such as rhythmic content analysis (Smith & Honing, 2008), feature extraction for music genre classification (Andén & Mallat, 2011; Grimaldi, Cunningham, & Kokaram, 2003; Tsunoo, Ono, & Sagayama, 2009; Tzanetakis*,* Essl, & Cook, 2001), pitch contour extraction and melodic indexing in "query-by-humming" systems (Jeon, Ma, & Ming Cheng, 2009; Jeon & Ma, 2011), denoising (Berger, Coifman, & Goldberg, 1994; Yu*,* Mallat, & Bacry, 2008) and audio compression (Dobson, Yang, Whitney, Smart, & Rigstaa, 1996; Srinivasan & Jamieson, 1998).

Wavelets exhibit similarities to many information-processing steps in the human brain and have been extensively used in modelling vision (see, e.g., Kay, Naselaris, Prenger, & Gallant, 2008; Zhang, Zhang, Huang, & Tian, 2005; Zhang, Shan, Qing, Chen, & Gao, 2009). In hearing, auditory perception in the cochlea and the auditory pathway has been modelled using bandpass filters based on the CWT and other wavelet-based techniques (Daubechies & Maes, 1996; Sinaga, Gunawan & Ambikairajah, 2003; Karmakar, Kumar & Patney, 2011). The interesting mathematical properties of wavelets and their applicability to modelling neural mechanisms motivate us to explore here the applicability of wavelets to the symbolic level of music description (i.e., to notes and their properties).

## 2.2 Symbolic music representation and analysis with wavelets

Although wavelets have been used extensively for analysing music audio, the use of the WT is scarce in the symbolic domain. One isolated example is Pinto's (2009) use of the DWT to index melodic sequences with few wavelet coefficients, obtaining improved retrieval results compared to the direct use of the melodies.

A Western staff-notation score depicts a piece of music as a set of notes, specifying (amongst other things) the pitch, relative onset time and relative duration of each note. In a MIDI file, the pitch of each note is specified by its MIDI note number, which represents its chromatic pitch (see Meredith, 2006, pp. 126–129). For the purpose of wavelet analysis, a melody can be represented as a 1D signal, called a *pitch signal*, that indicates the chromatic pitch (MIDI note number) of the melody at each tatum time-point. The pitch signal can then be transformed into coefficients at different scales using the WT. A similar representation using Fourier analysis has been shown by Schmuckler (1999) to capture relevant information for melodic similarity.

### 2.2.1 Melodic segmentation

Music unfolds over time. This characteristic is the most prominent difference between music and visual art, engaging our brains in a prediction-expectation game of events occurring over time (Huron, 2006; Levitin, 2006). We do not know how a piece will develop or end until it finishes. However, as the music unfolds, we constantly identify segments that start somewhere, develop and end. Finding coherent segments, or *groups*, at various different time scales is a basic, automatic aspect of music cognition (Lerdahl & Jackendoff, 1983).

Most theoretical work in music perception has concentrated on the perceived associations of events, based on grouping, adapting visual Gestalt principles of similarity and proximity to musical perception. These theories include Tenney and Polansky's (1980) theory of *temporal Gestalt-units*, Lerdahl and Jackendoff's (1983) grouping structure theory and the *Local Boundary Detection Model* (LBDM) of Cambouropoulos (1997, 2001), which sets local boundaries according to change and proximity rules. The rules in these models address both local changes and longer-term averages, so that representing melodic movements at different scales with wavelet filters, leading to different levels of localization on the time-axis, appears to be an appropriate approach for deriving group boundaries.

### 2.2.2 Relation to neural mechanisms

Recent neuroscientific imaging work based on EEG, fMRI and MEG provides evidence that musical structure constantly engages the brain in a game of prediction, expectation and reward, based on long-term memory and statistical regularities of coded features (Trainor & Zatorre, 2009). Moreover, it has been observed that brain activity increases transiently at musical movement boundaries, as well as other non-musical event boundaries, and it has been suggested that segmentation is thus an essential perceptual component, occurring simultaneously at multiple time-scales as an adaptive mechanism that integrates recent past information to improve predictions about the near future (Kurby & Zacks, 2008).

Perceptual boundary detection has been successfully modelled with wavelets. For example, Gabor wavelets have been used to model the early stages of the visual pathway (Kay *et al.*, 2008; Nixon & Aguado, 2012; Zhang *et al*., 2005; Zhang *et al*., 2009). It therefore seems reasonable to hypothesise that a similar wavelet-based approach might successfully be used to model group boundary perception in melodies.

### 2.2.3 Melodic theory

Huron (1996) proposes a reductionist approach to melodic classification, summarizing the contour of a folk song by its first and final pitches, along with an average of all the pitches in between. He demonstrates that folk songs have arc-like contours, with an inverted 'U' shape being the most common. In his study, a melody is classified into one of nine types, depending on whether it describes a trajectory that is ascending, descending, horizontal or a combination of these basic types.

In Schenkerian analysis (Brown, 2005; Forte & Gilbert, 1982; Schenker, 1935), the musical surface or *foreground* is recursively reduced to a *fundamental structure* (*Ursatz*) by removing notes of progressively greater structural importance. In a wavelet representation, small-scale structures that occur only in the foreground (e.g., ornaments) will be represented only in the small-scale coefficients; whereas the higher structural levels (corresponding loosely to the background or fundamental structure) will be represented by the coefficients at greater time scales. In this way, wavelets at different scales can be used to extract structure at what would correspond to different transformational levels (*Schichten*) in the Schenkerian approach.

It is possible to understand many musical works as having been generated by the reverse of this hierarchical reduction process—that is, by the successive elaboration of a fundamental structure with less structural notes, until the detailed foreground or musical surface emerges. Wavelet filters emphasise different temporal scales in a pitch signal, thus providing a tool to focus on and discover musical structure at a variety of different temporal scales.

## 3  Method

We investigate the effectiveness of the WT to represent relevant properties of melodies in segmentation and classification tasks. Our input data are sequences of notes, represented as pitch signals. To these we apply the CWT and obtain a time-scale representation for structural analysis in classification tasks. Figure 1 a) presents the score representation of a melodic fragment, Figure 1 b) is the 1D pitch signal that represents it, and Figure 1 c) is its CWT by Haar wavelet, in a scalogram plotting the absolute coefficients, using darker colours for smaller values and brighter colours for larger values.

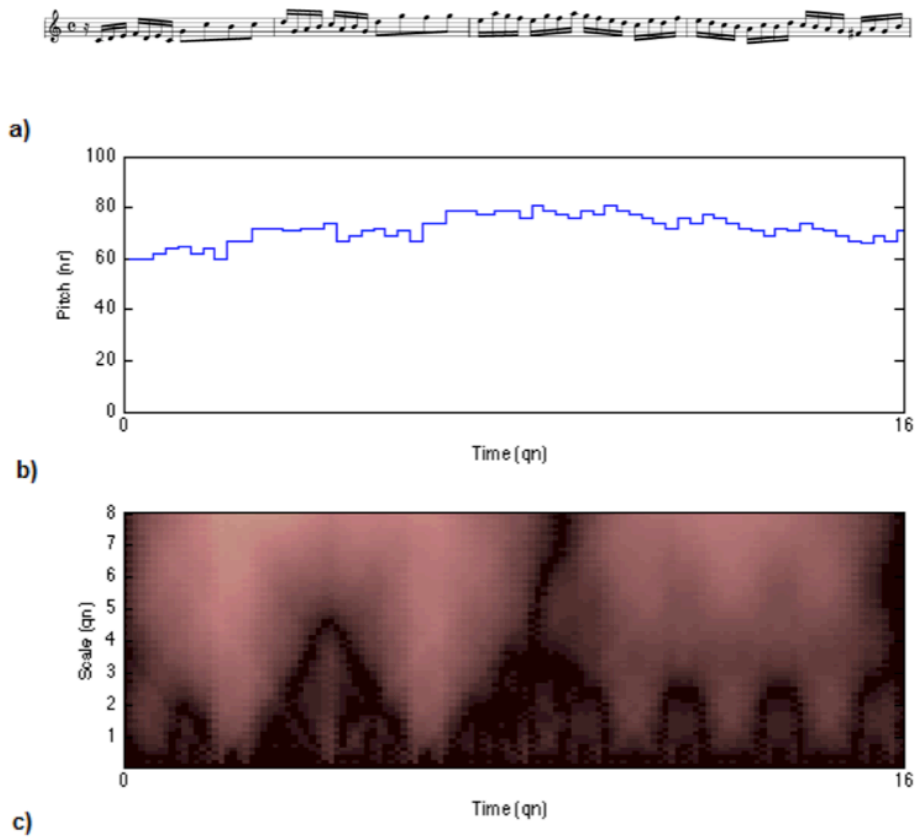

**Figure 1**. The opening bars of the upper part of J. S. Bach's *Invention* in C major (BWV 772), represented as a) a score, b) a pitch signal and c) a scalogram of the CWT (i.e., the absolute values of the coefficients).

### 3.1 Representation

We represent melodies as pitch signals or by the wavelet coefficients of the pitch signals.

#### 3.1.1 Pitch signal representation

A discrete *pitch signal v* with length $L$ is sampled from MIDI files at a rate $r$ in number of samples per quarter note (qn), so that we have a pitch value for every time point, expressed as $v[t]$. We use two different ways of treating rests: they are either *represented* by the value 0, or they are *removed* from the representation by the following procedure: if a rest occurs at the beginning of a sequence, it is replaced by the first pitch number that appears in the sequence, otherwise it is replaced with the pitch number of the note that immediately precedes it.

**Normalized pitch signal representation.** We normalize pitch signal segments by subtracting the average pitch in order to make the representation invariant to transposition. The normalization is applied after the segmentation.

### 3.1.2 Wavelet representation

The CWT[2] transforms a 1D signal into a set of coefficients $w_{s,u}$ using an analysing function $\psi_{s,u}(t)$, which is derived from the mother wavelet $\psi$ by scaling by a factor $s > 0$ and shifting in time by an amount $u \in \mathbb{R}$:

$$\psi_{s,u}(t) = \frac{1}{\sqrt{s}} \psi\left(\frac{t-u}{s}\right). \quad (1)$$

The coefficients $w_{s,u}$ are calculated for real valued wavelets as the inner product of the signal $v(t)$ and the analysing function $\psi_{s,u}(t)$:

$$w_{s,u} = \langle v, \psi_{s,u} \rangle = \int_{-\infty}^{+\infty} v(t) \psi_{s,u}(t) dt. \quad (2)$$

To avoid edge effects due to finite-length sequences (Torrence & Compo, 1998), we pad on both ends with a mirror image of the pitch signal (Woody & Brown, 2007). Once the coefficients are obtained, the segment that corresponds to the padding is removed, so that the signal maintains its original length.

We can treat coefficients on one scale as a function of the shift parameter with $w_s(u) = w_{s,u}$. Then the CWT acts as a *filter*, equivalent to the convolution of $v$ with the scaled and flipped real-valued wavelet. The CWT calculates the wavelet coefficients at all points $u$, so that the complete information of the pitch signal is still retained in the coefficients at one scale and it can be recovered using deconvolution, given a suitable wavelet.

For implementation on a computer, we can write equation (2) in a discretized version, where we compute the convolution for each translation $u$ and scale $s$:

$$w_s[u] = \sum_{l=1}^{L} \psi_{s,u}[l] v[l]. \quad (3)$$

---

[2] We follow the presentation by Antoine (1999). Signals processed by digital computers have to be discretized. The term "continuous" refers to the fact that all sample positions are used as shift values, as opposed to the discrete wavelet transform where shift values are much sparser.

The summation index only needs to run over the support of $\psi$, i.e., between the maximum and minimum time-points for which $\psi$ is not zero, which is typically considerably shorter than the signal *v*.

## 3.2 Wavelet choice

The selection of wavelet or analysing function depends on the kind of information that we want to extract from the signal, considering that the transform's coefficients combine information about the signal and the wavelet (Farge, 1992). The wavelet should give a compact representation of the variation in the signal that we are interested in. We use the Haar wavelet, which is defined by

$$\psi(t) = \begin{cases} 1, & if \quad 0 \le t < 1/2 \\ -1, & if \quad 1/2 \le t < 1 \\ 0, & otherwise \end{cases} \quad (4)$$

and has a shape as shown in Figure 2.[3]

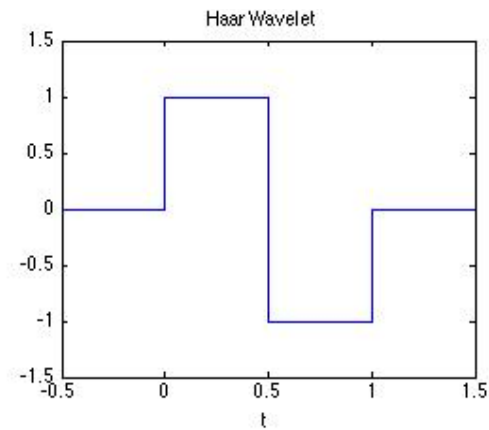

**Figure 2**. The shape of the Haar wavelet.

We selected the Haar wavelet because it matches the discontinuous, step-wise nature of the pitch signal. A continuous wavelet would require a combination of many small-scale components to represent the step transitions between pitches, obscuring the representation of pitch changes. On the other hand, the Haar wavelet is not suitable for continuous pitch data, which could represent vibrato, glissando, melismatic ornamentation, etc.

---

[3] The Haar function was introduced by Haar in 1910 (Haar, 1910). Equation (4) uses Mallat's (2009) notation.

The Haar wavelet has support on the time interval [0,$s$), and the inner product with the Haar wavelet calculates the difference between the averages of pitch in the first and second halves of that interval. In other words, the coefficient $w_{s,u}$ gives a measure of whether the melody is moving upwards or downwards over the scale period starting at position $u$.

Figure 3 illustrates the Haar wavelet shifted and scaled. In each of the three rows of sub-figures, different wavelet shifts can be seen (first vs. second column). The scale is 0.5 in the first row and 0.25 in the second and third rows.

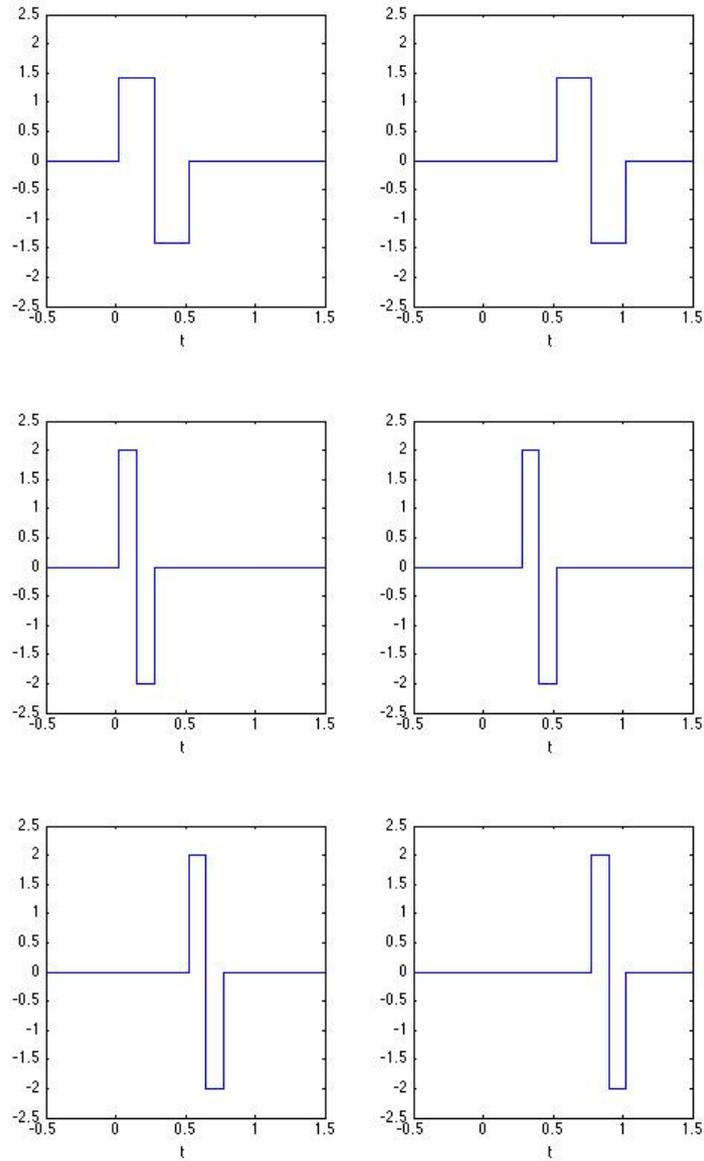

**Figure 3**. The Haar wavelet shifted and scaled.

## 3.3 Segmentation

We use the wavelet coefficients to determine melodic segments in two different ways, setting segmentation points either at local maxima or at zero crossings of the wavelet coefficients. Default segmentation points are set at the beginning and at the end of signals.

### 3.3.1 Zero crossing segmentation

Zero crossings occur when the inner product between the melody and the Haar wavelet is zero. This means that the average pitch in the first half of the scale period is equal to the average pitch in the second half of the scale period. See Figure 4 for illustration.

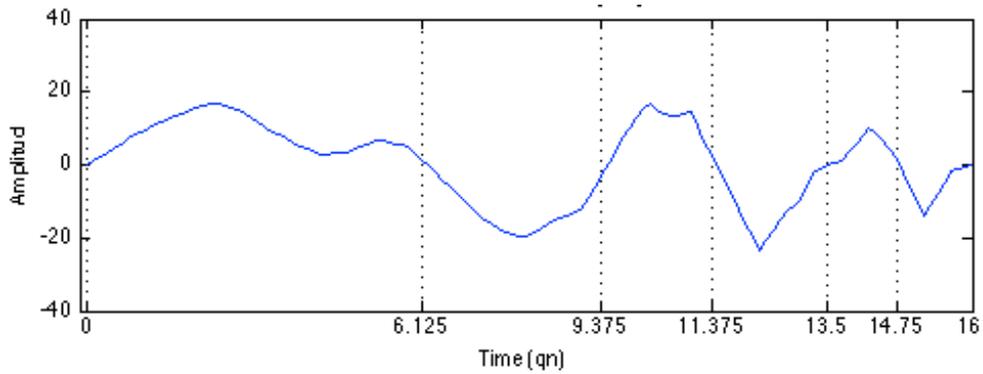

**Figure 4.** Wavelet coefficient signal at the scale of 4 for the first 16 qns of the sixth *Invention* in E major (BWV 777). Locations of zero-crossings are indicated by dotted vertical lines.

### 3.3.2 Local maxima segmentation

Local maxima in the wavelet representation occur when the shapes of the melody and the Haar wavelet correlate most. The inner product with the Haar wavelet of length $s$ can also be described as the difference of the average pitch during the first half of the wavelet minus the average pitch over the second half of the wavelet times $s$. Local maxima occur, therefore, where there is a locally maximal fall in average pitch content at the scale of the wavelet used. See Figure 5 for illustration.

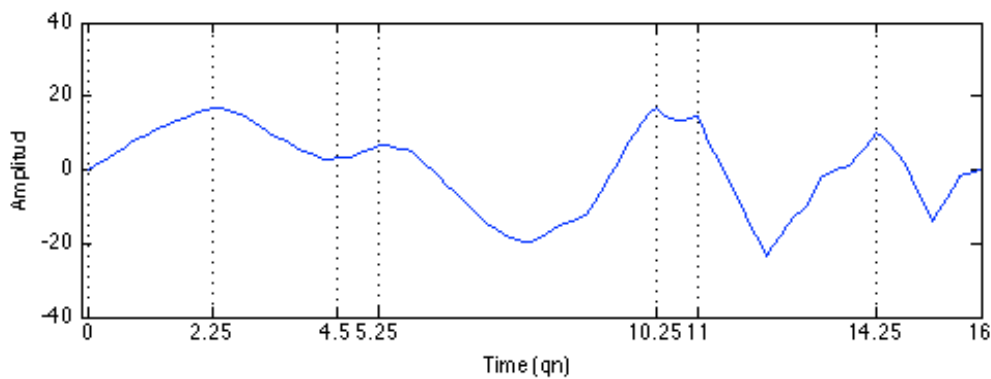

**Figure 5.** Wavelet coefficient signal at the scale of 4 for the first 16 qns of the sixth *Invention* in E major (BWV 777). Local maxima are indicated by dotted vertical lines.

### 3.3.3 Segment length normalization

In the evaluation tasks described below, the segments identified need to be classified, for which we introduce similarity measures on segments. We use the Euclidean and city-block distances, which entails that the segments need to be represented as vectors of equal length. However, segments are not generally of the same length when using the segmentation approaches described here. In order to obtain segments of equal length, we use two different procedures: we normalize the length of segments to the maximal segment length, or we define a maximal length for all segments and pad shorter segments as necessary with zeros at the end.

For comparison, we also segment using Eerola and Toiviainen's (2004) implementation of Cambouropoulos' (1997, 2001) LBDM (see above). The LBDM calculates a normalized boundary strength between 0 and 1 for the interval between each pair of consecutive notes in a melody (Cambouropoulos, 2001). In order to generate a specific segmentation, it therefore requires a threshold value between 0 and 1 to be defined.

### 3.4 Scale selection

In this study, we use the wavelet coefficients at only one scale, as we focus only on a single level of segmentation. By representing melodies by their wavelet coefficients at only one scale, we emphasise information on that time-scale in the signal, as discussed above. Small scales focus on short-term movements, while large scales emphasise the longer-term trend of the melody. We have tested dyadic multiples of quarter notes as scale values and selected those that yield the best classification results.

## 3.5 Classification

We use the wavelet representation and segmentation to perform classification of melodies with a *k*-Nearest-Neighbour (kNN) classifier. A kNN classifier is defined by a set of labelled items and a distance measure. It then assigns labels to a new item *x* by finding the *k* items that are closest to *x* according to the distance measure and choosing the label that occurs most often among these *k* items.

We use two different distance measures, *city-block* distance and *Euclidean* distance. The Euclidean distance between two segments, *st* and *sc*, is given by

$$d^E_{stsc} = \sqrt{\sum_{j=1}^{n}(st[j]-sc[j])^2}\ .$$

The city-block distance is given by

$$d^C_{stsc} = \sum_{j=1}^{n}|st[j]-sc[j]|.$$

## 3.6 Example

For illustration, Figure 6 presents an example of similarity measurements between a target segment (row 2) from a melody represented as pitch signal (row 1) and four test segments (rows 3 to 6). Test segment 3 has the smallest distance to the target segment when segment length normalization by zero padding is applied. On the other hand, if segment length normalization by interpolation is applied, the segment that has the smallest distance to the target segment is test segment 1.

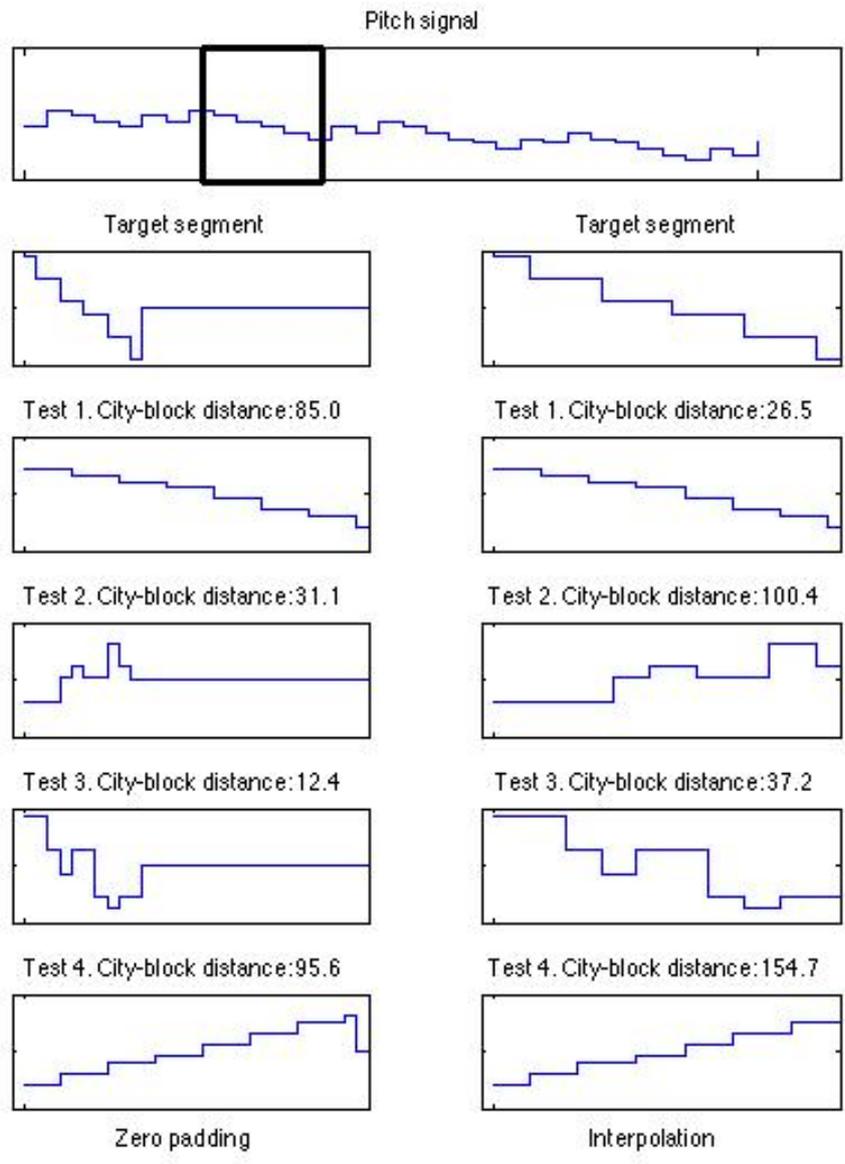

**Figure 6.** Illustration of a melodic segment (row 1) and similarity measurements between a target segment (row 2) and four test segments (rows 3 to 6). Segment length normalization by zero padding (left column) vs. segment length normalization by interpolation (right column). The black square in row 1 denotes the target segment.

# 4  Classification experiments

In this section, we present two experiments on different data sets[4]. One experiment is on recognizing the parent works of segments from Bach's *Two-Part Inventions* (BWV 772−786). The second experiment is on recognizing the tune families to which Dutch folk songs belong, using the Dutch Song Database (Grijp 2008; The Meertens Institute, 2012).

## 4.1  Experiment 1: Classification of segments from J. S. Bach's *Two-Part Inventions*

Music theorists describe J. S. Bach's *Inventions* as being coherently developed from a theme, the subject, that dominates each piece (see, e.g., Dreyfus, 1996). The *Invention's* subject is presented in the exposition, and it is contrapuntally treated across the (usually three) other sections (Stein, 1979). From this point of view, we hypothesize that the parent work of one of the later sections of an *Invention* can be successfully identified by finding the *Invention* with the exposition that the section resembles most closely in terms of melodic segments used.

For the 15 *Two-Part Inventions*, the classifier set $C$ is built from segments $sc_{i,j}$ from the expositions of all *Inventions*, where each segment can stem from either the upper or the lower part. $sc_{i,j}$ is the $j^{th}$ segment in *Invention i*. We define the length of the exposition as 16 qn, which is, of course, not accurate in all cases, but rather corresponds to the longest exposition in order to avoid including exposition material in the test sets possibly however, including material of the following section in the classifier. After the first 16 qn, each invention is divided into 3 sections of equal length to build the test sets. Each test set $T$ is built from segments $st$, where each $st$ can stem from either the upper or the lower part. We denote the $j^{th}$ segment in *Invention i* by $st_{i,j}$. To classify a segment $st$ to one of the 15 classes, we apply 1-NN classification. That is, we compute the distances between $st$ and all $sc$ in $C$, and classify $st$ to the class $i$ of the $sc_{i,j}$ that has the smallest distance to $st$. The section is assigned the class most frequently predicted by its segments. In both cases we use the next nearest point to break ties.

---

[4] The algorithms are implemented in MATLAB (R2012b, The Mathworks, Inc) using the Wavelet Toolbox and the MIDI Toolbox (Eerola & Toiviainen, 2004). We use the LBDM implementation of the MIDI Toolbox, and an update of Christine Smit's read_midi function (http://www.ee.columbia.edu/~csmit/matlab_midi.html, accessed 4 October 2012).

We test the classification accuracy of classifiers built from the first 4, 8 or 16 qn, on three, equally-divided sections after the exposition (see Figure 7), to study the development of the method's performance over the course of the *Invention*. We expect the classification rates to first decrease, reflecting the increasing degrees of variation of the original material and to increase towards the end, where the original material typically returns. We also compare different representations, segmentations and distance measures, as the performance can inform us about the suitability of these measures for representing the motivic coherence that music theorists describe in the *Inventions*.

We also test the effect of including contrapuntal variations in the classifier, because music theorists claim that these techniques are used for variation in the *Inventions* (and generally in imitative styles of music) (see, e.g., Dreyfus, 1996). Specifically, we considered inversion (reflection in a constant-pitch axis), retrograde (reflection in a constant-time axis) and retrograde inversion (rotation through a half turn) (see Figure 8). Contrapuntal variations are added as classes to the kNN classifier and we therefore have 4 times the number of classes.

We compare the wavelet representation with the normalized pitch signal representation, as described above. We evaluate the case when classifier and test sets contain one segment for each part and section, i.e. "*without segmentation*", and the case of applying a segmentation algorithm to create several segments from each part and section, which we call "*with segmentation*". We compare the results of zero crossing wavelet segmentation with two other segmentation methods: segmentation into segments of constant length, as a simplistic baseline segmentation, and segmentation with Cambouropoulos' LBDM as mentioned above. Local maxima wavelet segmentation was not used in this experiment as preliminary tests showed that segmenting at zero crossings produced better results in general for this dataset. Figure 9 shows, as an example, the first 16 qn of the upper voice of the first *Invention* (BWV 772) in the different combinations used for the experiments.

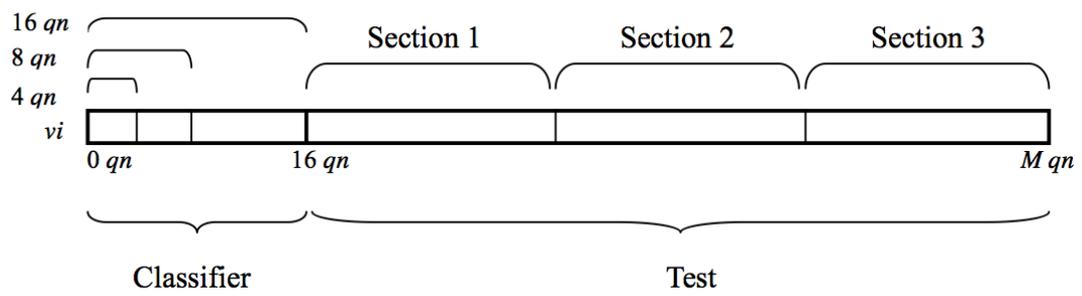

**Figure 7.** Scheme of classifier and test construction based on signal $v_i$.

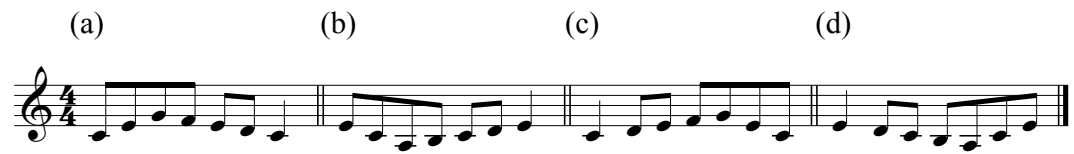

**Figure 8.** Contrapuntal variations: (a) prime form, (b) inversion, (c) retrograde and (d) retrograde inversion.

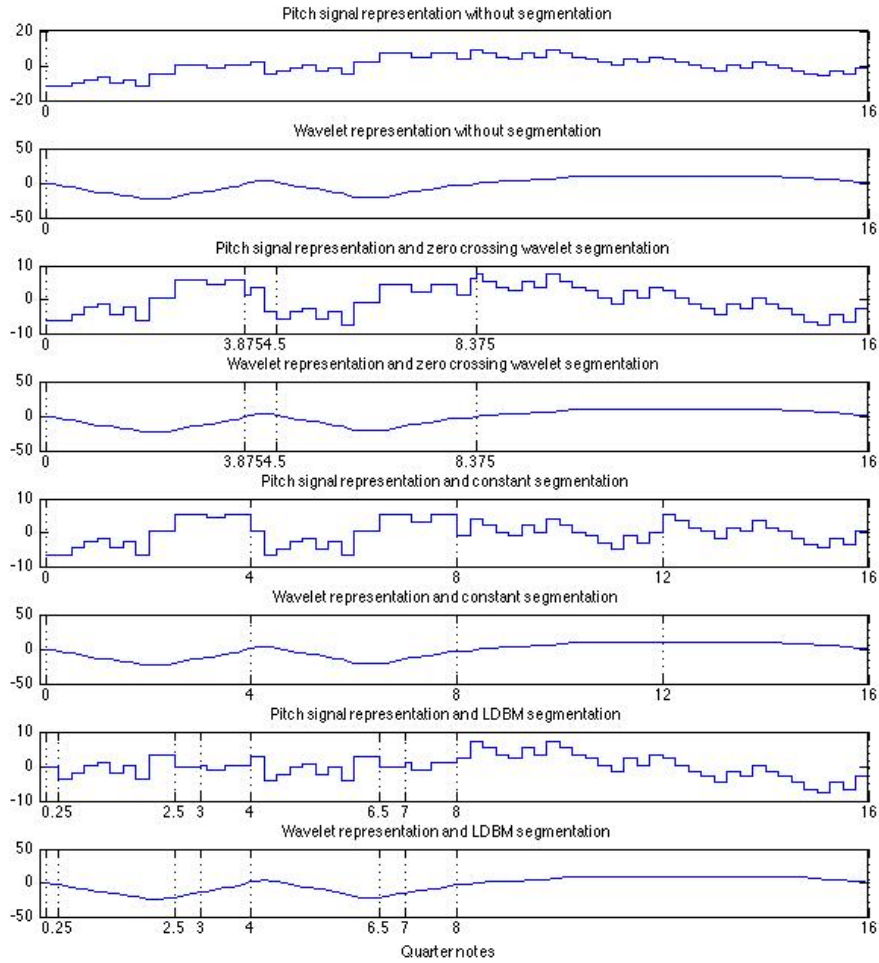

**Figure 9.** The first 16 qn of the upper voice of *Invention* 1 (BWV 772) in different combinations of representation and segmentation (the segmentation points are shown as vertical dotted lines): Normalized pitch signal representation (odd rows) and wavelet representation at scale of 4 qn (even rows), without segmentation (rows 1 and 2), wavelet segmentation at scale of 4 qn (rows 3 and 4), constant segmentation at 4 qn (rows 5 and 6), and LBDM with a threshold of 0.4 (rows 7 and 8). Pitch signal normalization takes places after segmentation, leading to pitch shifts between the original melody and the segments.

When the segments' lengths are normalized by zero padding, the length of segments is set to the maximal segment length, and shorter segments are padded as necessary with zeros at the end, even if they are segmented by constant length segmentation. In this case the sampling rate is not affected. When the segments' lengths are normalized by interpolation, the lengths of segments are resized to the maximal segment length by nearest neighbour interpolation (de Boor, 1978). This, of course, changes the sampling rate in most cases.

We used pitch signals initially sampled at 8 samples per quarter note (qn) and varied the following parameters to optimize classification performance:

- two melodic representations: normalized pitch signal representation (vr) and wavelet representation at scale of 1 qn (wr),
- without segmentation and with three segmentation methods: constant segmentation (cs) at 1 and 4 qn, LBDM with thresholds of 0.2 and 0.4 and zero crossing wavelet segmentation (ws) at scale 1 and 4 qn,
- segment length normalization by zero padding and by interpolation and
- Euclidean and city-block distance.

The optimal values of these parameters and the effect of representation, segmentation and contrapuntal variations will be presented in the results section.

## 4.2  Experiment 2: Classification of Dutch Folk Tunes

Folk tunes are a cultural heritage and interesting to study in the context of melodic classification because:

1) they present variation due to the process of oral transmission between generations;
2) understanding variations can help us understand cultural developments in music; and
3) there is a substantial body of research and data to support experiments and comparisons.

The Meertens Institute in Amsterdam hosts a collection of Dutch folk songs that has been digitized and classified into tune families according to similarity assessments done by experts (van Kranenburg, 2010). The Dutch Song Database we use contains 360 folk songs in 26 tune families, and is a subset of the collection known as "Onder de groene linde" (Grijp, 2008; The Meertens Institute, 2012). Automatic classification methods based on global features and string matching have been extensively tested by van Kranenburg (2010), and he concluded that recurrence of common motives is the most important musical factor in defining tune families.

For the Dutch tune family classification task, we designed two experiments, testing, among other parameters, the effect of segmentation. We use complete melodies or segments of melodies for classification.

### 4.2.1 Experiment 2-1: Classification without segmentation

In this experiment, we use complete melodies without segmentation. The songs of the Dutch Song Database are sampled to pitch signals of length $2^{10}$. We evaluate rest representation[5] and pitch normalization, as described in section 3.1. Moreover, we evaluate melodies as pitch signals or as wavelet coefficients. When melodies are represented as wavelet coefficients, we apply the CWT with Haar wavelet at a single scale. We evaluate classification accuracy with 1NN using city-block and Euclidean distances in leave-one-out cross validation on the corpus of 360 folk songs.

### 4.2.2 Experiment 2-2: Classification with segmentation

We build the classifier set *C* from all segments $sc_j$ of the whole corpus minus one—that is 359 labeled songs. The remaining song is used for testing. We use *kNN* classification, where *k*=1 to 5. We thus compute the distances between a test segment $st_j$ and all segments in *C*, and assign the segment to the most frequent class of the *k* segments with the smallest distances and the tune to the most frequent class of its segments. We calculate the classifiers' accuracies using all segments of all songs belonging to a tune family with 1 to 5 nearest neighbours and with two distance measures (Euclidean and city-block) in leave-one-out cross validation on the corpus of 360 folk songs.

In this second experiment with segmentation, we use once again the two types of melodic representations (normalized pitch signal and wavelet coefficients at one scale) but only two segmentation models: LBDM and local maxima of wavelet coefficients. Zero crossings were not used in this experiment as preliminary tests showed that segmenting at local maxima produced better results in general for this dataset[6]. The MIDI files of this collection are initially sampled at 8 samples per qn. We apply the CWT with the Haar wavelet using a dyadic set of 8 scales. Melodies are represented as normalized pitch signals (vr) or as the resulting wavelet coefficients (wr). Signals are segmented by the wavelet coefficients' local maxima (ws), or by the local boundary detection model LBDM using thresholds from 0.1 to 0.8 in steps of 0.1. We explore the parameter space with a grid search, testing all combinations of representations and segmentations: wavelet representation (wr), normalized pitch signal representation (vr), wavelet segmentation (ws), LBDM (LBDM) segmentation. Segment length normalization is done by zero padding and by interpolation.

---

[5] We also tested the way that rests are represented in normalized pitch signals by assigning the value zero to rests, subtracting the average pitch (excluding rests) and assigning the value zero to rests again after normalization. This practice produced worse results than the way that rests are represented in the normalized pitch signal representation described in section 3.1.

[6] We ran some tests with segmentation points at local extrema (i.e., local minima and maxima), but, in general, results with local maxima were better.

# 5 Results and discussion

## 5.1 Results of experiment 1: Classification of segments from J. S. Bach's *Two-Part Inventions*

### 5.1.1 Experiment 1-1. Classification without segmentation

Table 1 shows the best accuracies with a corpus of the 15 *Two-Part Inventions* by J. S. Bach (BWV 772−786) without segmentation. The parameters used to achieve the values shown in Table 1 are:

- pitch signals sampled at 8 samples per qn,
- normalized pitch signal representation,
- wavelet representation at the scale of 1 qn,
- 1-nearest neighbour classifier with city-block or Euclidean distance, and
- length normalization by zero padding or by interpolation.

|  | City-block | | Euclidean | |
|---|---|---|---|---|
|  | (wr) | (vr) | (wr) | (vr) |
| Mean NC | 0.1778 | 0.0889 | 0.1333 | 0.0889 |
| Std-Dev. NC | 0.0385 | 0.0770 | 0.0667 | 0.1018 |
| Mean CP | 0.1333 | 0.1556 | 0.0667 | 0.1333 |
| Std-Dev. CP | 0.0667 | 0.1388 | 0.0000 | 0.1155 |

**Table 1.** Experiment without segmentation. Summary of the best classification accuracies over three sections of the inventions, mean and standard deviation (Std-Dev.) of the classifiers build from the first 16 qn. Classifier built from the exposition (NC), and the classifier built from the exposition and its contrapuntal variations (CP). Combinations: wavelet representation (wr), normalized pitch signal representation (vr)..Appendix A, Table A3 shows the results of all combinations tested in the experiment.

This approach is a baseline experiment, which does not use segment information or alignment, and the observed accuracies are above chance level (6.66%) but very low as expected.

### 5.1.2 Experiment 1-2. Classification with segmentation

For this corpus and experiment, segmentation improves the classification rates substantially. Figures 10 and 11 show the classification performance on each section, the effect of segmentation and representation (rows vs. columns), the effect of including contrapuntal techniques (Figure 10 vs. Figure 11) and the number of quarter notes used for the classifiers (red, green and blue lines). The remaining fixed parameter values were chosen such that the best results were achieved in the majority of the cases shown (Appendix A, Tables A1 and A2 summarize the results of all other parameterisations). The used parameter values are:

- normalized pitch signal representation,
- wavelet representation at the scale of 1 qn,
- zero crossing wavelet segmentation at the scale of 1 qn,
- LBDM segmentation at a threshold of 0.2,
- constant segmentation at 1 qn,
- 1-nearest neighbour classifier with city-block distance, and
- segment length normalization by zero padding.

The classification results vary widely, with segmentation method having a stronger effect than representation type. Wavelet segmentation combined with wavelet representation produces the best classification results when using 16 quarter notes of the exposition.

Including contrapuntal variations is clearly detrimental when using wavelet segmentation and to some degree when using LBDM, but improves performance with constant segmentation. This result was unexpected, as a common view in musicology is that inversion, retrograde and retrograde inversion are important principles of variation in J. S. Bach's inventions (e.g. Stein, 1979) and would therefore help in recognising the inventions. However, the lower-than-expected recognition rates achieved with our contrapuntal variation classifier may be due to the fact that we use chromatic pitch representations rather than ones based on diatonic (or "morphetic") pitch (see Meredith, 2006, pp. 126–9).

The classification performance generally decreases from the 1st to the 2nd sections and it rises from the 2nd to the 3rd sections, to some degree conforming to the expectation of increased similarity between the final section and the exposition.

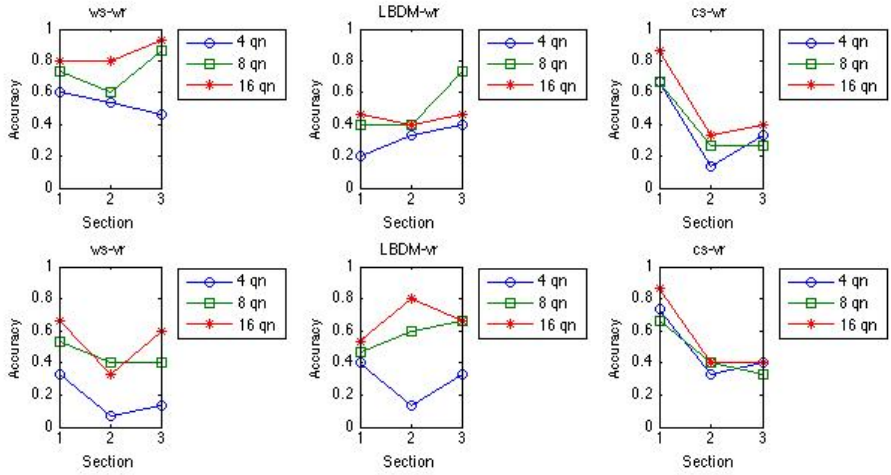

**Figure 10.** Performance for each section with the classifier based on the exposition.

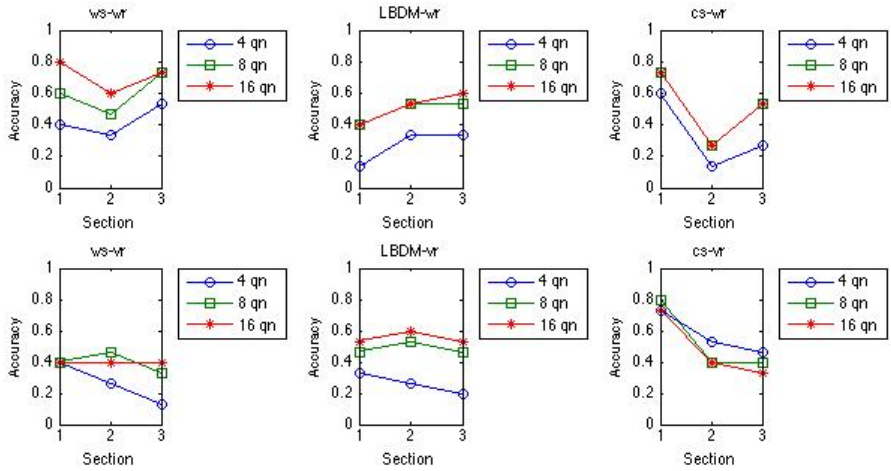

**Figure 11.** Performance for each section with the classifier based on the exposition and its contrapuntal variations.

## 5.2 Results of experiment 2: Classification of Dutch Folk Tunes

### 5.2.1 Classification without segmentation

Table 2.1 shows the classification rates obtained in the experiment on the corpus of 360 Dutch Folk songs without segmentation, using complete melodies. The parameter values are:

- pitch signals of length $2^{10}$,
- normalized pitch signal representation,
- wavelet representations at a single scale and
- classification in leave-one-out cross validation with 1 nearest neighbours using Euclidean and city-block distances.

|      | City-block | Euclidean | City-block | Euclidean |
|------|------------|-----------|------------|-----------|
|      | Rests removed |        | Rests represented by zeros | |
| (vr) | 0.8806     | 0.8694    | 0.7944     | 0.7056    |
| (wr) | 0.8556     | 0.8306    | 0.7472     | 0.7222    |

**Table 2**. Classification accuracy observed for different methods. Pitch signal representation (vr) and wavelet representation (wr) combined with different distance measures and rest treatment.

For this experiment, removing rests from the representation produced better classification accuracies. We therefore removed rests from the representation for the experiment with segmentation. The use of complete melodies represented as pitch signals without filtering produces the best results.

### 5.2.2 Classification with segmentation

Contrary to the effect seen in experiment 1, segmentation did not produce a significant change in the classification rates, even varying several parameters. Figure 12 shows the classification rates obtained with segmentation, where brighter colours indicate higher rates. The parameter values are:

- pitch signals initially sampled at 8 samples per qn,
- normalized pitch signal representation,
- wavelet representations using a dyadic set of 8 scales,
- local maxima wavelet segmentation using a dyadic set of 8 scales,
- LBDM segmentation using thresholds from 0.1 to 0.8 in steps of 0.1,
- classification with 1 to 5 nearest neighbours using city-block distances, and
- segment length normalization by zero padding.

Table 3 summarizes the best and worst classification rates with the parameters mentioned above. The effect of using segment length normalization by interpolation produces slightly lower results than segment length normalization by zero padding (see Table 4).

The results show that wavelet filtering of the melodic segments can improve classification performance compared to using the pitch signal directly. When segmentation is used, wavelet representation proves to be more discriminative than pitch signals independently of the segmentation method. The classification performance varies, obtaining best results at small representation scales and poor results at large scales, with the exception of the largest scale, which recovers its performance to some extent (see Figure 12).

In terms of segmentation, we observe that shorter segments produce better results when used with wavelet representation. This is contrary to the results of the LBDM applied to pitch signals, where shorter segments produce worse results than larger ones. We observe an improvement towards threshold 0.4 and a gradual improvement towards the threshold of 0.8, which corresponds to larger segments, meaning that using the complete melodic sequences or a combination of complete melodies and melodic segments can lead to better classification results. Indeed, as shown in the first part of this second experiment using the Dutch Song Database, the classification rates improve when using complete melodies represented as pitch signals.

In general, the city-block distance performs slightly better than Euclidean distance and the wavelet representation works better than the normalized pitch signal representation. In addition, we studied the effect of using more than one nearest neighbour. It can be observed that using one and two nearest neighbours produced the best results. Different effects are seen when using values greater than 2 for $k$ in the $k$NN, but in general the performance decreases as $k$ increases.

The best classification rates are achieved by using the wavelet representation and segmentation using 1 or 2 nearest neighbours at small scales. This suggests that the melodies in this corpus contain typically several similar segments that are typical for that family. This agrees with van Kranenburg's (2010) claim that recurrent motives are important for determining the family of a folk song in the Dutch Song Database. On the other hand, the results of van Kranenburg *et al*. (2013) using string-matching are considerably better, suggesting that information on the order of the segments also plays an important role.

| | | City-block distance | | | | |
|---|---|---|---|---|---|---|
| represent.-segment. | Value | Nearest Neighbours | | | | |
| | | 1 | 2 | 3 | 4 | 5 |
| wr-ws | best | **0.8556** | **0.8556** | 0.8333 | 0.8306 | 0.7972 |
| | worst | 0.4833 | 0.4833 | 0.4639 | 0.45 | 0.4167 |
| wr-LBDM | best | 0.8417 | 0.8417 | 0.8083 | 0.8028 | 0.7778 |
| | worst | 0.4417 | 0.4417 | 0.4556 | 0.4417 | 0.4139 |
| vr-ws | best | 0.8139 | 0.8139 | 0.7972 | 0.7778 | 0.7472 |
| | worst | 0.5194 | 0.5194 | 0.5194 | 0.5139 | 0.5583 |
| vr-LBDM | best | 0.7889 | 0.7889 | 0.7778 | 0.75 | 0.725 |
| | worst | 0.4139 | 0.4139 | 0.3861 | 0.3778 | 0.3806 |

| | | Euclidean distance | | | | |
|---|---|---|---|---|---|---|
| represent.-segment. | Value | Nearest Neighbours | | | | |
| | | 1 | 2 | 3 | 4 | 5 |
| wr-ws | best | **0.8417** | **0.8417** | 0.8306 | 0.8194 | 0.7917 |
| | worst | 0.4667 | 0.4667 | 0.4583 | 0.4333 | 0.4167 |
| wr-LBDM | best | 0.8111 | 0.8111 | 0.8083 | 0.7889 | 0.7694 |
| | worst | 0.4472 | 0.4472 | 0.4528 | 0.4333 | 0.4139 |
| vr-ws | best | 0.8083 | 0.8083 | 0.7806 | 0.7667 | 0.7444 |
| | worst | 0.5194 | 0.5194 | 0.5333 | 0.525 | 0.5639 |
| vr-LBDM | best | 0.7778 | 0.7778 | 0.7444 | 0.7333 | 0.7083 |
| | worst | 0.4111 | 0.4111 | 0.3722 | 0.3806 | 0.3806 |

**Table 3.** Summary of the accuracies for the combinations: wavelet representation and wavelet segmentation (wr-ws), wavelet representation and local boundary detection model (wr-LBDM), pitch signal representation and wavelet segmentation (vr-ws), pitch signal representation and local boundary detection model (vr-LBDM), segment length normalization by zero padding.

| represent.-segment. | Value | Euclidean distance ||||| 
| | | Nearest Neighbours ||||| 
| | | 1 | 2 | 3 | 4 | 5 |
| --- | --- | --- | --- | --- | --- | --- |
| wr-ws | best | **0.8361** | **0.8361** | 0.8194 | 0.7944 | 0.7722 |
| | worst | 0.4306 | 0.4306 | 0.3944 | 0.3889 | 0.3722 |
| wr-LBDM | best | 0.8056 | 0.8056 | 0.7972 | 0.7611 | 0.7389 |
| | worst | 0.3611 | 0.3611 | 0.3528 | 0.3306 | 0.3083 |
| vr-ws | best | 0.7833 | 0.7833 | 0.7694 | 0.7806 | 0.7556 |
| | worst | 0.5111 | 0.5111 | 0.5444 | 0.5167 | 0.5000 |
| vr-LBDM | best | 0.7833 | 0.7833 | 0.7639 | 0.7667 | 0.7500 |
| | worst | 0.3917 | 0.3917 | 0.3667 | 0.3639 | 0.3611 |

| represent.-segment. | Value | City-block distance ||||| 
| | | Nearest Neighbours ||||| 
| | | 1 | 2 | 3 | 4 | 5 |
| --- | --- | --- | --- | --- | --- | --- |
| wr-ws | best | **0.8472** | **0.8472** | 0.8333 | 0.8111 | 0.7944 |
| | worst | 0.4306 | 0.4306 | 0.4222 | 0.4000 | 0.3556 |
| wr-LBDM | best | 0.8139 | 0.8139 | 0.7917 | 0.7806 | 0.7528 |
| | worst | 0.3306 | 0.3306 | 0.3333 | 0.3083 | 0.3083 |
| vr-ws | best | 0.7944 | 0.7944 | 0.7861 | 0.7917 | 0.7583 |
| | worst | 0.5083 | 0.5083 | 0.5389 | 0.5306 | 0.5583 |
| vr-LBDM | best | 0.8028 | 0.8028 | 0.7833 | 0.7806 | 0.7528 |
| | worst | 0.4028 | 0.4028 | 0.3722 | 0.3639 | 0.3694 |

**Table 4.** Summary of the accuracies for the combinations: wavelet representation and wavelet segmentation (wr-ws), wavelet representation and local boundary detection model (wr-LBDM), pitch signal representation and wavelet segmentation (vr-ws), pitch signal representation and local boundary detection model (vr-LBDM), segment length normalization by interpolation.

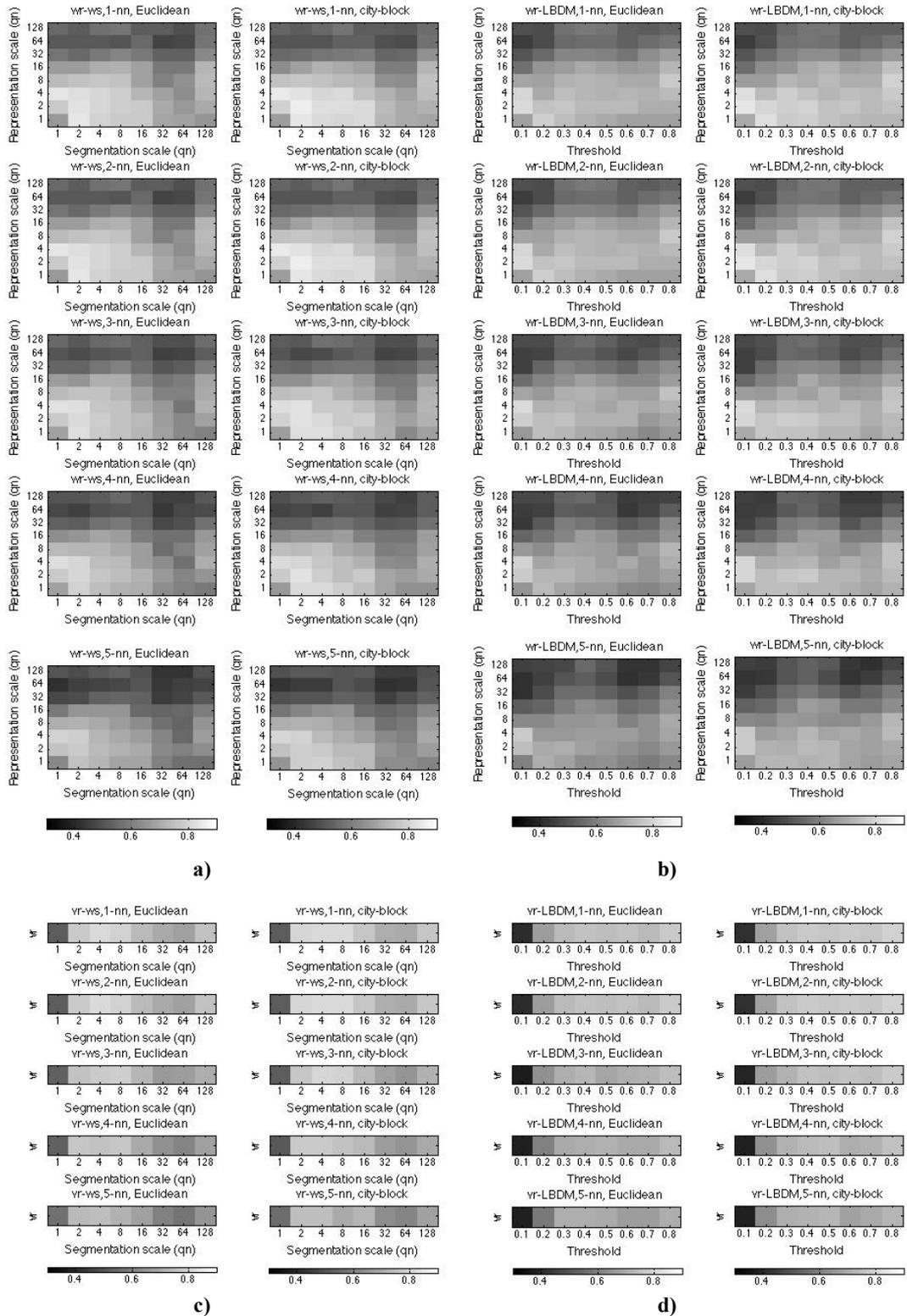

**Figure 12.** Accuracies for the combinations: a) wavelet representation (wr) and wavelet segmentation (ws), b) wavelet representation (wr) and local boundary detection model (LBDM), c) pitch signal representation (vr) and wavelet segmentation (ws), pitch signal representation (vr) and local boundary detection model (LBDM), segment length normalization by zero padding.

## 5.3 Discussion

We have presented two experiments, in which continuous Haar-wavelet filtering was applied in two musicologically motivated classification tasks. The results of the first experiment support the view that there are strong, intra-opus, motivic relations within Bach's *Two-Part Inventions* that allow for the parent works of sections from these pieces to be identified, depending on the amount of material used from the exposition, along with the approaches used to segment and represent the music. The negative effect of adding contrapuntal variations in the classifiers in connection with wavelet segmentation is interesting and may suggest that the similarities captured by wavelets are different to and in some way incompatible with contrapuntal variations we have used in the experiment. On the other hand, this effect could also be an artefact of the specific type of pitch representation used—we intend to explore this further in future work.

When the wavelet-based approach was used to identify the tune families of songs in a database of Dutch folk songs, it proved to work slightly better than using the LBDM with direct melody comparison and slightly worse than using complete melodies without filtering. However, results with string-matching methods reported by van Kranenburg *et al.* (2013) are considerably better. This indicates that the overall sequential structure of the melody is relevant for this task, which is ignored in the segmentation approach. This is supported by the observation that the wavelet-based classifier performs similarly at small and large scales, with different $k$ values and for different distance metrics, indicating that the relevant information may not be just in the segments.

Segment length normalization by zero padding produces slightly better results than normalization by interpolation. This suggests that the structure of segments is related to their length and the effect of zero padding does not negatively influence the reliability of similarity measurement.

Melodic segmentation has a different effect between the two experiments possibly due to musical differences between the Dutch folk tunes and Bach's *Inventions* or due to different principles determining whether two tunes should be in the same tune family or whether two melodic excerpts belong in the same piece.

# 6 Conclusion

In this paper, we have presented a method for using wavelets to represent and segment melodies for classification and we have evaluated it on two different musicological classification tasks. Our main contribution has been to introduce and demonstrate the potential of a novel, wavelet-based approach to modelling melodic structure.

The results of the experiments reported here suggest that a method employing a wavelet-based approach to representing and segmenting the data can out-perform one that uses a direct pitch-time representation and Gestalt-based or constant-duration segmentation in the task of predicting which work in a collection contains a given query segment. When the task was to identify the musicologically defined tune family to which a given folk song belongs, our wavelet-based approach worked only slightly better than one based on Gestalt principles and slightly worse than one without segmentation using pitch melodies. However, it was clearly out-performed by string-matching methods, which is probably due to the fact that, in this task, the overall structure of the compared melodies contains relevant information that our classification method is not using, regardless of whether or not wavelets are used.

We propose that the positive results of wavelet representation and segmentation can be understood by viewing the wavelets in terms of the pitch trend over the scale duration. Focusing on an appropriate time-scale, giving less weight to short-term movement as well as the average pitch (i.e., transposition), can make relevant parts of the melodic contour more prominent in the distance measure.

# 7 Future work

There are several further aspects of modelling melodic perception with wavelets that have not been explored in this study, including the problem of automatic scale determination, and the relation between musical style and features in wavelet coefficient representations.

Understanding the wavelet analysis better in terms of musical properties may help improve the results for melodic similarity. Multiple scales could be used for hierarchical segmentation. Using a selective combination of scales and exploring metrical information derived from songs' periodicities could be used to develop a method for scale selection. Applying machine learning to develop more complex wavelet-based feature extraction from melodies could also be a very interesting way to use the wavelet representation on symbolic music data.

We also aim to identify the cognitive mechanisms that underlie the effectiveness of the wavelet-filtering approach and to explain why coefficient zero-crossings work better in some classification tasks while coefficient local maxima work better in others.

We generally aim in future research to gain a deeper understanding of the musical meaning and perceptual relevance of wavelet-based music representation and segmentation.

Appendix A

|   |   |   | Wavelet rep. (wr) | | | Pitch signal rep. (vr) | | |
|---|---|---|---|---|---|---|---|---|
|   |   |   | (ws) | (LBDM) | (cs) | (ws) | (LBDM) | (cs) |
| A | Rests represented | Mean NC | 0.8444 | 0.4444 | 0.5333 | 0.5333 | 0.6667 | 0.5556 |
|   |   | Std-Dev. NC | 0.0770 | 0.0385 | 0.2906 | 0.1764 | 0.1333 | 0.2694 |
|   |   | Mean CP | 0.7111 | 0.5111 | 0.5111 | 0.4000 | 0.5556 | 0.4889 |
|   |   | Std-Dev. CP | 0.1018 | 0.1018 | 0.2341 | 0.0000 | 0.0385 | 0.2143 |
|   | Rests removed | Mean NC | 0.7556 | 0.6000 | 0.5333 | 0.4000 | 0.6222 | 0.4889 |
|   |   | Std-Dev. NC | 0.0770 | 0.0667 | 0.2309 | 0.1333 | 0.0770 | 0.2776 |
|   |   | Mean CP | 0.5778 | 0.6000 | 0.5556 | 0.2667 | 0.4889 | 0.4000 |
|   |   | Std-Dev. CP | 0.2037 | 0.0667 | 0.2694 | 0.1155 | 0.0385 | 0.2309 |
| B | Rests represented | Mean NC | 0.4889 | 0.4889 | 0.3556 | 0.5556 | 0.3556 | 0.3556 |
|   |   | Std-Dev. NC | 0.1018 | 0.0770 | 0.2694 | 0.0385 | 0.1018 | 0.2776 |
|   |   | Mean CP | 0.3556 | 0.4000 | 0.3556 | 0.4222 | 0.3333 | 0.4000 |
|   |   | Std-Dev. CP | 0.1018 | 0.0667 | 0.2694 | 0.1018 | 0.0667 | 0.2000 |
|   | Rests removed | Mean NC | 0.4000 | 0.6000 | 0.3556 | 0.3111 | 0.3778 | 0.4222 |
|   |   | Std-Dev. NC | 0.0000 | 0.0000 | 0.3289 | 0.0770 | 0.1018 | 0.3289 |
|   |   | Mean CP | 0.4000 | 0.5111 | 0.3778 | 0.3778 | 0.4000 | 0.3778 |
|   |   | Std-Dev. CP | 0.0667 | 0.1678 | 0.3791 | 0.1388 | 0.1155 | 0.1925 |
|   |   |   | LN-Zero padding | | | | | |
|   |   |   | Wavelet rep. (wr) | | | Pitch signal rep. (vr) | | |
|   |   |   | (ws) | (LBDM) | (cs) | (ws) | (LBDM) | (cs) |
| A | Rests represented | Mean NC | 0.8000 | 0.4444 | 0.5778 | 0.6444 | 0.4444 | 0.5778 |
|   |   | Std-Dev. NC | 0.0667 | 0.0385 | 0.3151 | 0.0385 | 0.1678 | 0.3079 |
|   |   | Mean CP | 0.6889 | 0.3778 | 0.5778 | 0.3556 | 0.4667 | 0.5111 |
|   |   | Std-Dev. CP | 0.0385 | 0.0385 | 0.2524 | 0.1018 | 0.0667 | 0.2694 |
|   | Rests removed | Mean NC | 0.8000 | 0.5333 | 0.5556 | 0.3556 | 0.4667 | 0.4667 |
|   |   | Std-Dev. NC | 0.0667 | 0.0000 | 0.3421 | 0.1678 | 0.1333 | 0.2906 |
|   |   | Mean CP | 0.6444 | 0.4444 | 0.5778 | 0.2222 | 0.4667 | 0.4000 |
|   |   | Std-Dev. CP | 0.1018 | 0.0385 | 0.2694 | 0.1388 | 0.0667 | 0.2906 |
| B | Rests represented | Mean NC | 0.3556 | 0.1778 | 0.3556 | 0.4444 | 0.1778 | 0.3778 |
|   |   | Std-Dev. NC | 0.1018 | 0.1388 | 0.2694 | 0.0770 | 0.1540 | 0.2037 |
|   |   | Mean CP | 0.3778 | 0.2000 | 0.3778 | 0.4889 | 0.1778 | 0.4222 |

|   |   |   | Wavelet rep. (wr) | | | Pitch signal rep. (vr) | | |
|   |   |   | (ws) | (LBDM) | (cs) | (ws) | (LBDM) | (cs) |
|---|---|---|---|---|---|---|---|---|
|   |   | Std-Dev. CP | 0.0385 | 0.1333 | 0.1925 | 0.0385 | 0.0385 | 0.2037 |
|   |   | Mean NC | 0.4000 | 0.2667 | 0.3333 | 0.2889 | 0.2222 | 0.4000 |
|   |   | Std-Dev. NC | 0.1333 | 0.1764 | 0.2906 | 0.1018 | 0.1018 | 0.2000 |
|   |   | Mean CP | 0.3111 | 0.1778 | 0.4222 | 0.3111 | 0.2222 | 0.3333 |
|   | Rests removed | Std-Dev. CP | 0.1388 | 0.1018 | 0.2694 | 0.1018 | 0.1018 | 0.2309 |
|   | LN-Interpolation | | | | | | | |

Table A1. Classification accuracies over three sections of the inventions, mean and standard deviation (Std-Dev.) values of the classifiers build from the first 16 qn, using only the exposition (NC) and the exposition and its contrapuntal variations (CP) for wavelet representation at the scale of 1 qn and normalized pitch signal representation using city-block distance and 1NN. (A) corresponds to segmentation: (ws) at 1 qn, (LBDM) at threshold 0.2 and (cs) at 1 qn. (B) corresponds to segmentation (ws) at 4 qn, (LBDM) at threshold 0.4 and (cs) at 4 qn.

|   |   |   | Wavelet rep. (wr) | | | Pitch signal rep. (vr) | | |
|   |   |   | (ws) | (LBDM) | (cs) | (ws) | (LBDM) | (cs) |
|---|---|---|---|---|---|---|---|---|
| A | Rests represented | Mean NC | 0.8667 | 0.4667 | 0.5333 | 0.5111 | 0.6222 | 0.6444 |
|   |   | Std-Dev. NC | 0.0667 | 0.1333 | 0.2906 | 0.1678 | 0.0385 | 0.2524 |
|   |   | Mean CP | 0.7111 | 0.4444 | 0.5111 | 0.3778 | 0.6000 | 0.5333 |
|   |   | Std-Dev. CP | 0.0385 | 0.0770 | 0.1678 | 0.0385 | 0.0000 | 0.2404 |
|   | Rests removed | Mean NC | 0.7333 | 0.5333 | 0.5111 | 0.4000 | 0.6000 | 0.5556 |
|   |   | Std-Dev. NC | 0.0667 | 0.0000 | 0.2524 | 0.1333 | 0.0667 | 0.2341 |
|   |   | Mean CP | 0.6000 | 0.5778 | 0.5778 | 0.2444 | 0.4444 | 0.4444 |
|   |   | Std-Dev. CP | 0.2000 | 0.0385 | 0.2341 | 0.1018 | 0.0385 | 0.2524 |
| B | Rests represented | Mean NC | 0.4667 | 0.4444 | 0.3333 | 0.3556 | 0.3778 | 0.4222 |
|   |   | Std-Dev. NC | 0.1155 | 0.1388 | 0.2309 | 0.1018 | 0.1018 | 0.2341 |
|   |   | Mean CP | 0.3778 | 0.3333 | 0.4444 | 0.3778 | 0.3556 | 0.4222 |
|   |   | Std-Dev. CP | 0.1678 | 0.0000 | 0.3079 | 0.1678 | 0.0385 | 0.2143 |
|   | Rests removed | Mean NC | 0.3778 | 0.4222 | 0.3333 | 0.3333 | 0.4000 | 0.3556 |
|   |   | Std-Dev. NC | 0.1018 | 0.1388 | 0.3528 | 0.2000 | 0.1333 | 0.3289 |
|   |   | Mean CP | 0.3778 | 0.4000 | 0.3333 | 0.2667 | 0.3778 | 0.3556 |
|   |   | Std-Dev. CP | 0.1388 | 0.1155 | 0.3528 | 0.1764 | 0.0770 | 0.2694 |
|   | LN-Zero padding | | | | | | | |

|   |   |   | Wavelet rep. (wr) | | | Pitch signal rep. (vr) | | |
|   |   |   | (ws) | (LBDM) | (cs) | (ws) | (LBDM) | (cs) |
|---|---|---|---|---|---|---|---|---|
| A | Rests represented | Mean NC | 0.7778 | 0.4667 | 0.4889 | 0.6000 | 0.4444 | 0.6222 |
|   |   | Std-Dev. NC | 0.1018 | 0.0667 | 0.3289 | 0.0667 | 0.1018 | 0.2776 |
|   |   | Mean CP | 0.7111 | 0.3778 | 0.5778 | 0.3556 | 0.4444 | 0.4889 |
|   |   | Std-Dev. CP | 0.0385 | 0.0385 | 0.2037 | 0.0770 | 0.0770 | 0.2143 |
|   | Rests removed | Mean NC | 0.8000 | 0.4667 | 0.4889 | 0.3778 | 0.4667 | 0.5778 |
|   |   | Std-Dev. NC | 0.0667 | 0.1155 | 0.3421 | 0.1018 | 0.1333 | 0.3079 |
|   |   | Mean CP | 0.6667 | 0.4000 | 0.5333 | 0.2222 | 0.4000 | 0.4000 |
|   |   | Std-Dev. CP | 0.1155 | 0.0667 | 0.3055 | 0.1388 | 0.1155 | 0.2906 |
| | | Mean NC | 0.3333 | 0.2222 | 0.4667 | 0.4222 | 0.2222 | 0.3556 |
| | | Std-Dev. NC | 0.0667 | 0.1018 | 0.2667 | 0.0385 | 0.0770 | 0.2143 |
| | | Mean CP | 0.3333 | 0.2889 | 0.4222 | 0.3778 | 0.2000 | 0.4444 |
| | Rests represented | Std-Dev. CP | 0.0000 | 0.1540 | 0.2776 | 0.0770 | 0.0667 | 0.1388 |
| | | Mean NC | 0.3778 | 0.2222 | 0.3556 | 0.3111 | 0.3111 | 0.3333 |
| | | Std-Dev. NC | 0.1018 | 0.1678 | 0.3906 | 0.1388 | 0.0770 | 0.2404 |
| | | Mean CP | 0.3333 | 0.1778 | 0.2889 | 0.2000 | 0.2667 | 0.3778 |
| B | Rests removed | Std-Dev. CP | 0.0667 | 0.0770 | 0.3289 | 0.1333 | 0.1764 | 0.2037 |
| | | LN-Interpolation | | | | | | |

Table A2. Classification accuracies over three sections of the inventions, mean and standard deviation (Std-Dev.) values of the classifiers build from the first 16 qn, using only the exposition (NC) and the exposition and its contrapuntal variations (CP) for wavelet representation at the scale of 1 qn and normalized pitch signal representation using Euclidean distance and 1NN. (A) corresponds to segmentation: (ws) at 1 qn, (LBDM) at threshold 0.2 and (cs) at 1 qn. (B) corresponds to segmentation (ws) at 4 qn, (LBDM) at threshold 0.4 and (cs) at 4 qn.

| | | | City-block | | Euclidean | |
|---|---|---|---|---|---|---|
| | | | (wr) | (vr) | (wr) | (vr) |
| | | Mean NC | 0.1778 | 0.0889 | 0.1333 | 0.0889 |
| | | Std-Dev. NC | 0.0385 | 0.0770 | 0.0667 | 0.1018 |
| | | Mean CP | 0.1333 | 0.1556 | 0.0667 | 0.1333 |
| | Rests represented | Std-Dev. CP | 0.0667 | 0.1388 | 0.0000 | 0.1155 |
| | | Mean NC | 0.1333 | 0.0667 | 0.0667 | 0.1111 |
| | | Std-Dev. NC | 0.1155 | 0.0000 | 0.0000 | 0.0385 |
| | | Mean CP | 0.1556 | 0.1111 | 0.1333 | 0.0667 |
| P | Rests removed | Std-Dev. CP | 0.0385 | 0.1388 | 0.0000 | 0.0667 |
| | | Mean NC | 0.0667 | 0.0889 | 0.0889 | 0.0444 |
| | | Std-Dev. NC | 0.0667 | 0.0385 | 0.0770 | 0.0385 |
| | | Mean CP | 0.0889 | 0.1333 | 0.1111 | 0.0667 |
| | Rests represented | Std-Dev. CP | 0.0770 | 0.0667 | 0.1018 | 0.0667 |
| | | Mean NC | 0.0222 | 0.0667 | 0.0222 | 0.0667 |
| | | Std-Dev. NC | 0.0385 | 0.0000 | 0.0385 | 0.0000 |
| | | Mean CP | 0.0222 | 0.1556 | 0.0444 | 0.0889 |
| I | Rests removed | Std-Dev. CP | 0.0385 | 0.1018 | 0.0385 | 0.0385 |

Table A3. Classification accuracies without segmentation over three sections of the inventions, mean and standard deviation (Std-Dev.) values of the classifiers build from the first 16 qn, using only the exposition (NC) and the exposition and its contrapuntal variations (CP) for wavelet representation at the scale of 1 qn (wr) and normalized pitch signal representation (vr) using city-block and Euclidean distances and 1nn. (P) corresponds to length normalization by zero padding. (I) corresponds to interpolation length normalization by interpolation.